\pdfoutput=1

\documentclass[11pt]{article}

\usepackage{emnlp2021}

\usepackage{times}
\usepackage{latexsym}

\usepackage[T1]{fontenc}

\usepackage[utf8]{inputenc}

\usepackage{microtype}

\usepackage{booktabs}
\usepackage{multirow}
\usepackage{multicol}
\usepackage{amsmath}

\usepackage{graphicx}

\title{Does Pretraining for Summarization Require Knowledge Transfer?}

\author{Kundan Krishna, Jeffrey Bigham, Zachary C. Lipton\\
  Carnegie Mellon University \\
  5000 Forbes Avenue \\
  Pittsburgh, PA \\
  \texttt{\{kundank,jbigham,zlipton\}@andrew.cmu.edu} \\
  }

\begin{document}
\maketitle
\begin{abstract}
Pretraining techniques leveraging enormous datasets
have driven recent advances in text summarization.
While folk explanations suggest 
that \emph{knowledge transfer} accounts 
for pretraining's benefits,
little is known about why it works 
or what makes a pretraining task or dataset suitable.
In this paper, we challenge the \emph{knowledge transfer} story,
showing that pretraining on documents consisting
of character n-grams selected at random,
we can nearly match the performance
of models pretrained on real corpora.
This work holds the promise 
of eliminating upstream corpora, 
which may alleviate some concerns 
over offensive language, bias, and copyright issues.
To see whether the small residual benefit of using real data
could be accounted for by the structure of the pretraining task,
we design several tasks motivated 
by a qualitative study of summarization corpora.
However, these tasks confer no appreciable benefit,
leaving open the possibility of a small role
for knowledge transfer.\footnote{The code and the datasets used in the paper are
available at \url{https://github.com/acmi-lab/pretraining-with-nonsense}}

\end{abstract}

\section{Introduction}
Despite the widespread success of pretrained models 
when fine-tuned on diverse downstream NLP tasks,
such as summarization~\cite{qi2020prophetnet, raffel2020exploring}, 
question answering, sentiment analysis etc~\cite{yang2019xlnet}, 
scientific explanations for these benefits remain unknown.
Several works have claimed 
that pretrained models learn 
linguistic knowledge from the pretraining corpus
\citep{lina2019open,tenney2019bert,manning2020emergent},
leading to a popular, but unproven hypothesis
that credits knowledge transfer 
for the improvements seen on downstream tasks.
However, several recent findings 
test the plausibility of this account.
For example, benefits of pretraining have been observed
even when the upstream text has 
no syntactic structure~\citep{sinha2021masked} 
and others have shown benefits even
when the upstream corpus is from 
a different domain entirely,
such as music~\citep{papadimitriou2020learning} 
or amino acid sequences \citep{chiang2020pre}

In this work, we show that, surprisingly,
pretraining objectives previously demonstrated 
to be helpful for summarization~\citep{zou2020pre},
continue to deliver significant benefits
even when applied on text consisting of
randomly sampled nonsense words.
Because the text consists of nonsense words
sampled independently and uniformly,
it seems difficult to fathom a credible argument 
that the synthetic corpus encodes 
linguistic knowledge in any relevant sense.
Nevertheless, when pretraining transformer-based 
sequence-to-sequence models using this nonsense text,
we achieve significant performance boosts
on multiple downstream summarization benchmarks
that nearly match the performance of pretrained transformers.

Remarkably, when pretraining with synthetic tasks,
using real data offers no benefit over the nonsense data, 
on multiple summarization benchmarks. 
Thus, we investigate whether a pretraining task
better aligned with the demands of summarization
might close this residual gap.
We design a collection of pretraining tasks
inspired by some of the basic primitive operations
that appear to be common routines
required in order to create real-world summaries.
We carried out an extensive survey of
public summarization datasets 
spanning different domains,
and catalogued several elementary operations 
that were frequently invoked
in producing summaries
(e.g., extract sentences on a specific topic, 
or determine the most frequent 
among a set of relevant terms). 
In our proposed pretraining corpus, 
the summary is created by carrying out these
elementary operations on the input.
However, we find that our pretraining tasks 
deliver comparable performance gains 
to those proposed in ~\citet{zou2020pre}
leaving the small gap open.
On CNN-Dailymail and Rotowire benchmarks, 
where median summary lengths are 
73 and 456 tokens respectively, 
using our pretraining tasks with nonsense text
results in achieving on average $95\%$ of the performance gain in ROUGE-1
that standard T5 pretrained models enjoy
relative to randomly initialized T5. 
By contrast, on XSum and Rottentomatoes, 
where summaries are shorter (29 and 32 tokens respectively), 
we realize a relatively modest $37\%$ of the benefit on average.

The takeaways from our results are two-fold:
First, these results challenge
our understanding of why pretraining helps in summarization, 
suggesting that a large portion of the benefits seen may 
not be due to any knowledge transfer, 
but simply better initialization from an 
optimization perspective. 
Second, the ability to realize 
the benefits of pretraining
without using real-world data 
could alleviate concerns 
regarding bias, offensive speech, 
and intellectual property 
associated with using web-scale pretraining 
corpora of unknown provenance
\citep{davidson2019racial, bordia2019identifying}.

\section{Related Work}
Recently, multiple pretrained models 
have shown remarkable performance on text summarization. 
These models have been pretrained on real data
with diverse denoising tasks,
including masked language modeling~\cite{raffel2020exploring}, 
text infilling~\cite{zhang2020pegasus},
and sentence reordering~\cite{lewis2020bart}, among others.
While these pretraining objectives
have shown benefits across multiple NLP tasks, 
\citet{zou2020pre} proposed a set of three denoising pretraining tasks 
that are specifically motivated by summarization
and deliver performance comparable
to previous pretrained models. 
Our paper shows that the pretraining tasks in \citet{zou2020pre} 
improve summarization performance 
even if the pretraining corpus is artificial 
and does not encode any linguistic structure.

Our work extends a growing body of scientific literature 
that questions commonly-held beliefs 
about what properties of a pretraining corpus 
lead to improvements on different downstream tasks. 
Recently, \citet{sinha2021masked} showed 
that word order in pretraining documents 
has negligible impact on downstream performance 
on the GLUE benchmark. 
Even pretraining on sequences 
from different modalities such as 
Java code and amino acid sequences \citep{chiang2020pre}
have shown benefits on GLUE benchmark,
Similarly, for the task of language modeling,
pretraining on musical scores, or even
artificial sequences of nested parentheses
has shown to achieve better perplexity on 
a human language~\citep{papadimitriou2020learning}.
Our results go further---here the source documents 
contain no natural data at all,
nor do they exhibit any non-trivial structure.

Recently, some machine learning theory literature 
has begun to question the mechanism
by which transfer learning works. 
For example, \citet{neyshabur2020being} attribute 
the benefits to low-level statistics of the data 
and optimization considerations rather than feature reuse. 
In other related work, 
\citet{maennel2020neural} show that networks 
pretrained on randomly labeled data 
sometimes enjoy considerable performance improvements on downstream tasks. 

\section{Generating the Nonsense Corpus} 
For generating the nonsense pretraining corpus,
we use an artificial vocabulary
to create base documents
that has little resemblance to any real language. 
Our vocabulary simply consists of the first 5000 
3-letter character combinations 
using the English alphabet in lexical order starting
from the right (\textit{aaa, baa, caa, ..., aab, bab, ...}). 
Each sentence is generated 
by sampling each word in it independently 
from the uniform distribution over the entire vocabulary, 
and ending it with a period 
(see Figure~\ref{fig:pretraining_pipeline}
for a sample nonsense document).
The length of each sentence is selected 
uniformly from 5 to 15 words.
The number of sentences per document is selected according 
to the pretraining task that it is used for.
For the pretraining tasks 
proposed in \citet{zou2020pre}, we sample sentences
until the document reaches 512 tokens in length.
For our pretraining tasks (introduced later),
number of sentences in a document is decided 
by sampling uniformly from 7 to 13 sentences.

\section{STEP Pretraining Tasks}

STEP pretraining tasks are a collection of 3 tasks defined by ~\citet{zou2020pre}.
Next Sentence Generation (NSG) 
provides the first half of a document as input 
and the target is to generate the latter half. 
Sentence Reordering (SR) presents a document
with its sentences shuffled in random order, 
and requires generating the original document 
with correct sentence order. 
Masked Document Generation (MDG) masks out
a contiguous sequence of tokens in the base document 
and requires generating the original document 
while correctly filling-in the masked tokens. 
More details and hyperparameters 
can be found in the original paper.
\section{Our Pretraining Tasks}
To develop our pretraining tasks,
we first undertook a qualitative analysis 
of existing summarization datasets.
We surveyed all summarization papers 
published in the last 10 years 
with more than 25 citations, 
cataloguing a list of the summarization datasets 
that were used in them.
We observed that datasets 
can be grouped together 
according to domain 
(e.g., news and conversations).
We grouped the 28 retrieved datasets 
into 14 domains
(see the Appendix, Table~\ref{tab:dasasets_domains}). 
We selected a single dataset from each domain 
to analyze what summaries consist of 
and what \emph{skills} their creation requires.

From each selected dataset,
we manually inspected ten randomly sampled input-summary pairs,
looking for primitive subtasks 
that seem to express \emph{skills} 
(informally) that are required 
in order to create the summaries
demanded by this dataset
for at least two of the ten instances.
Since we need to create artificial 
input-summary pairs for each subtask,
we only chose subtasks for which it was possible
to create large number of such artificial pairs.
For example, in the Samsum dataset~\citep{gliwa2019samsum} 
which requires summarizing conversations between people, 
a frequently necessary subtask is 
to infer the unfolding social scenario
(e.g. a fight, or a person helping another)
but it is difficult to create a large number 
of varied artificial conversations 
that reflect the situation.
On the other hand, subtasks
such as extracting those sentences 
that address some specific topic,
or (even simpler) extracting 
the first sentence of the input
are simple enough to facilitate 
creating data points programatically.
Note that while copying the first sentences
might seem like a trivial or uninteresting pretraining task, it can be very useful.
For example, in news summarization datasets
the lead-3 baseline (copying over first 3 sentences as the summary) 
works very well \citep{brandow1995automatic, grenander2019countering}.

Based on this analysis, we 
developed 21 elementary tasks,
including copying specific content,
performing numerical operations, and more.
See Table~\ref{tab:21tasks} for 
full details on the slate of tasks.


\begin{table*}[]
    \centering
    \resizebox{\linewidth}{!}{
    \begin{tabular}{c|c}
    \toprule
    \textbf{Elementary subtask} & \textbf{Description} \\
    \midrule
CheckKeyword &	Check if the input has a special keyword or not. \\
\hline
ClassifyKeyword &	Output the category of keyword occurring in the input \\
\hline
MajorityKeyword &	Out of two given keywords, find which one occurs more number of times \\
\hline
CopyFirstSentence &	Copy first sentence \\
\hline
CopyBulleted &	Copy over a bullet point (sentence starting with a bullet marker). \\
\hline
CopyQuoted &	Copy text within quotes \\
\hline
CopyLastSentence &	Copy last sentence \\
\hline
CopyKwdOneSentence &	Copy the sentence that contains a keyword  \\
\hline
CopyKwdMultipleSentInOrder &	Copy all sentences containing any keyword in their order of appearance. \\
\hline
CopyKwdMultipleSentSorted &	Copy all sentences containing any keyword, sorted by the keywords \\
\hline
CopyKwdMultipleSentShuffled &	Copy all sentences containing keywords in any  order.  \\
\hline
ReplaceClassKeyword &	Replace an object's mention with its category (e.g. apple $\xrightarrow{}$ fruit) \\
\hline
CompareNumbers &	Given two numbers in the text, say which one is bigger \\
\hline
SumOfNumbers &	Sum all numbers in the input \\
\hline
ThresholdNumber &	Check if a number in the input is above a threshold  \\
\hline
LargestNumber &	Find out largest of one or more numbers in the input. \\
\hline
TruncateSentence &	Copy a sentence but only till the cutoff keyword is encountered \\
\hline
BreakClauses &	Break a single sentence into multiple ones containing one clause each \\
\hline
JoinClauses &	Join clauses from multiple sentences to make one longer sentence \\
\hline
ParaphraseWords &	Copy a sentence while replacing its keywords with one of its synonyms \\
\hline
TopicSegregation &	Copy sentences containing keywords from different classes into separate sections \\
\bottomrule
    \end{tabular}
    }
    \caption{21 extracted elementary summarization subtasks and their descriptions (detailed version is in Appendix)}
    \label{tab:21tasks}
\end{table*}


\vspace{5px}
\noindent 
\textbf{Generating artificial summaries \quad} 
To create an input-summary pair 
using an elementary task 
from Table~\ref{tab:21tasks},
we first create a  base document and then
(when required by the task)
modify it by adding the requisite keywords.
For example, \emph{CopyKwdOneSentence} 
uses a keyword to mark the sentence to copy. 
The keywords added for tasks are also meaningless
like \textit{keyword1, keyword2}. 
Then the corresponding elementary operation is applied 
to generate the summary from this modified input.

\begin{figure}[t]
    \includegraphics[width=0.5\textwidth]{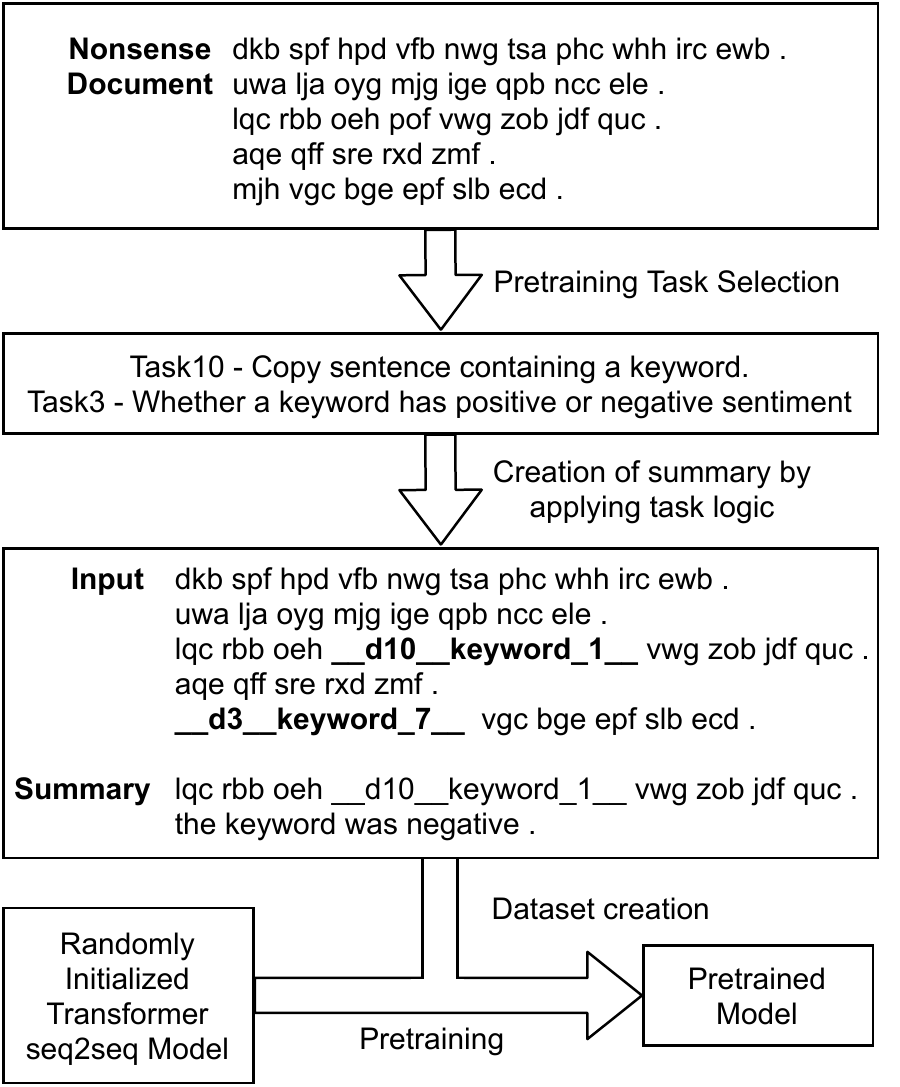}
    \caption{Procedure to create pretraining dataset using the nonsense corpus and our proposed pretraining tasks}
    \label{fig:pretraining_pipeline}
\end{figure}

The pretraining dataset that we create
involves multiple elementary operations
in each input-summary pair.
To create the input-summary pair from a nonsense document, 
we first sample 3 elementary tasks
and sequentially modify the input as needed by each task. 
Then, we generate the summary sentence(s)
as required for each elementary task
and concatenate them to constitute the overall summary. 
Here, the different keywords added to the input 
signal to the model which tasks are 
required to generate the summary.
The procedure is illustrated in Figure~\ref{fig:pretraining_pipeline}.

\section{Summarization Benchmarks}
We fine-tune and evaluate our models 
on 4 downstream summarization benchmarks.

\vspace{5px}
\noindent \textbf{CNN-Dailymail-10K}~\citep{see2017get}\quad
Contains news articles and summaries 
from CNN and Dailymail websites. 
We use only $10k$ instances for training 
(randomly sampled from the training set) 
so that the impact of pretraining is more visible. 
However, we still evaluate the fine-tuned model 
on the full test set.

\vspace{5px}
\noindent \textbf{XSum-10K}~\citep{Narayan2018DontGM}\quad
Also a news summarization dataset.
Again, we train on a random subset 
of $10k$ instances from the training set. 

\vspace{5px}
\noindent \textbf{Rottentomatoes}~\citep{wang2016neural}\quad 
This dataset concerns summarizing 
critical reviews of movies
found on the website \texttt{rottentomatoes.com}. 

\vspace{5px}
\noindent \textbf{Rotowire}~\citep{wiseman2017challenges} \quad
Here, the task is to process the box-score of a basketball game 
(often requiring numerical reasoning) 
to create a post-game summary. 

\begin{table*}[h!]
    \centering
    \resizebox{\linewidth}{!}{    
    \begin{tabular}{l | c c c | c c c | c c  c | c c c}
    \toprule
    \textbf{Model}     &  \multicolumn{3}{c}{\textbf{CNN-DM-10K}}  &  \multicolumn{3}{c}{\textbf{XSum-10K}}  &  \multicolumn{3}{c}{\textbf{Rotten Tomatoes}} &  \multicolumn{3}{c}{\textbf{Rotowire}}  \\
\toprule
{} &  R1 &  R2 &  RL &  R1 &  R2 &  RL &  R1 &  R2 &  RL &  R1 &  R2 &  RL \\
\midrule
T5-OffShelf                     &     39.38 &     18.08 &     27.71 &    29.18 &     8.69 &    22.62 &              24.73 &               9.00 &              19.64 &        37.50 &        12.85 &        19.85 \\
T5-RI                   &      9.86 &      1.06 &      7.49 &    15.49 &     2.48 &    12.76 &              10.17 &               0.41 &               8.66 &         4.02 &         0.72 &         3.68 \\
\midrule
\multicolumn{13}{c}{{Nonsense Upstream Corpus}}\\
\midrule
T5-OurTasks                     &     35.23 &     14.77 &     24.03 &    20.36 &     4.15 &    16.23 &              15.72 &               2.06 &              12.51 &        39.10 &        11.81 &        19.94 \\
T5-STEPTasks                    &     35.78 &     14.98 &     23.60 &    21.49 &     4.56 &    16.78 &              13.22 &               0.88 &              10.83 &        29.82 &         7.45 &        16.74 \\
\midrule
T5-STEPTask-NSG                 &      9.20 &      0.80 &      7.19 &    15.78 &     2.24 &    12.44 &              12.31 &               0.71 &              10.60 &        33.65 &         7.60 &        17.90 \\
T5-STEPTask-SR                  &     28.63 &     10.67 &     20.35 &    21.47 &     4.70 &    16.62 &              10.89 &               0.51 &               9.18 &        25.68 &         5.39 &        15.29 \\
T5-STEPTask-SR-adjusted         &      7.24 &      0.63 &      5.69 &    15.04 &     2.00 &    12.12 &              11.18 &               0.46 &               9.51 &        20.00 &         2.74 &        12.08 \\
T5-STEPTask-MDG                 &     34.50 &     14.45 &     23.77 &    20.76 &     4.13 &    16.45 &              11.78 &               0.70 &               9.89 &        36.22 &        10.53 &        18.73 \\
T5-STEPTask-MDG-adjusted &     10.15 &      0.93 &      7.78 &    16.12 &     2.20 &    13.09 &              15.07 &               1.38 &              11.69 &          20.39 &          3.77 &          11.97 \\
\midrule
\multicolumn{13}{c}{{Real Upstream Corpus}}\\
\midrule
T5-OurTasks                &     34.06 &     13.88 &     23.21 &    22.27 &     5.09 &    17.60 &              19.16 &               5.26 &              15.65 &        38.57 &        11.89 &        19.68 \\
T5-STEPTasks               &     32.04 &     12.93 &     22.55 &    23.37 &     5.68 &    18.42 &              20.89 &               6.29 &              17.05 &        37.63 &        10.89 &        19.57 \\
\midrule
\multicolumn{13}{c}{{PG Models Randomly Initialized vs Pretrained (Nonsense Upstream Corpus)}}\\
\midrule
PG-RI                   &     29.68 &     11.75 &     21.82 &    17.66 &     3.57 &    14.62 &              19.63 &               6.43 &              16.62 &        30.61 &         8.66 &        17.74 \\
PG-OurTasks                     &     29.82 &     11.78 &     21.91 &    16.81 &     3.43 &    13.95 &              19.02 &               6.57 &              16.38 &        26.94 &         6.81 &        16.77 \\
PG-STEPTasks                    &     29.44 &     11.74 &     21.67 &    17.65 &     3.54 &    14.55 &              17.70 &               5.89 &              15.34 &          31.16 &          8.49 &          17.85 \\
\bottomrule
\end{tabular}
}
    \caption{Rouge scores achieved by different models on four summarization benchmarks.}
    \label{tab:main_rouge_results}
\end{table*}

\section{Experiments and Results}

First, we pretrain the transformer-based 
sequence-to-sequence architecture 
used by the T5 model~\cite{raffel2020exploring}, 
on different corpora, 
each containing $100k$ input-summary pairs
to get different pretrained models. 
We use the T5-small architecture  
in all experiments.
Next, we fine-tune each model 
on the downstream tasks
and measure performance via ROUGE score
(Table~\ref{tab:main_rouge_results}).
We also present the models' performance 
on next token prediction in summaries using
accuracy and log-likelihood in the Appendix 
(Table~\ref{tab:tokpred_results}).
To frame the comparison, 
we include the performance of
the official \textbf{T5} model 
and of a randomly initialized model
using the same architecture (\textbf{T5-RI}) .

Pretraining with either our proposed pretraining tasks (\textbf{OurTasks}), 
or STEP tasks (\textbf{STEPTasks})
performs much better than random initialization, 
even when using nonsense data.
For all summarization benchmarks except RottenTomatoes,
the performance remained comparable 
when we used real upstream data from Wikipedia
to create the pretraining datasets.
This suggests that for some summarization benchmarks, 
there might be little or no additional benefit 
provided by using real world pretraining text. 

Looking at individual STEPTasks, NSG has no training signal 
since the output is completely independent of the input,
but surprisingly it leads to improvements in Rotowire benchmark.
SR and MDG performed much better than NSG on CNN-DM and XSum, 
likely because they involve copying sentences/unmasked tokens from the input.
We created \emph{adjusted} versions of these pretraining datasets,
where there was no copying needed 
and it led to a drop in performance 
on both pretraining tasks, 
bringing it close to \textbf{T5-RI} for CNN-DM and XSum. 
In SR-adjusted, the task is to output only the numerical order
in which sentences should be copied
(versus actually generating the full output).
In MDG-adjusted, the task is to only output the masked-out tokens 
(versus outputting the entire document, including unmasked tokens).

A randomly initialized pointer-generator model~\citep{see2017get} (\textbf{PG-RI})
performs far better than a randomly initialized T5 model.
However, T5-architecture models pretrained on nonsense text
were able to outperform pointer-generator on 3 out of 4 benchmarks, 
suggesting that transformer models pretrained on nonsense text can be 
a better choice than using non-pretrained LSTM based models.
Interestingly, pretraining the PG model on either \textbf{OurTasks} or \textbf{STEPTasks} 
did not lead to any additional improvement.

Models pretrained separately on each task from \textbf{OurTasks}
exhibit strong differences in their performance on
CNN-Dailymail-10K benchmark (Table~\ref{tab:individual_task_effect}). 
Models pretrained on \emph{TopicSegregation} and \emph{CopyKwdMultipleSent-Shuffled} 
outperform others significantly. 
The two worst performing models
were pretrained on \emph{CompareNumbers} and \emph{SumOfNumbers}, 
and these models were unable to perform any better 
than random guessing on the pretraining task itself.
By contrast, most other pretrained models 
were able to solve their pretraining task correctly
more than $99\%$ of times 
(see Table~\ref{tab:individual_task_effect_full} 
in Appendix for full details).

\begin{table}[]
    \centering
    \resizebox{\linewidth}{!}{
    \begin{tabular}{l|cc|c}
    \toprule
    \textbf{Pretraining task} &    \textbf{R1}  & \textbf{R2} & \textbf{Pr\%}\\
    \midrule
TopicSegregation                     &   23.04 &    7.79  &  99.90 \\
CopyKwdMultipleSent-Shuffled         &   23.34 &    5.46  &  99.66 \\
TruncateSentence                     &   17.07 &    2.50  &  1.00 \\
\midrule
LargestNumber                        &    6.52 &    0.58  &  99.88 \\
SumOfNumbers                         &    5.03 &    0.40  &  25.06 \\
CompareNumbers                       &    1.89 &    0.04  &  48.88 \\
    \bottomrule
    \end{tabular}
    }
    \caption{The 3 best and worst performing pretraining tasks 
    according to performance of their pretrained models
    on CNN-Dailymail-10K (R1,R2), 
    and their accuracy on the pretraining task (Pr\%).}
    \vspace{-10px}
    \label{tab:individual_task_effect}
\end{table}

\section{Conclusion}
This paper demonstrated that transformer models 
pretrained on randomly generated nonsense data 
deliver remarkable performance gains 
across multiple summarization tasks, 
compared to their randomly initialized version. 
This suggests that a substantial part
of the observed benefits of pretraining 
can not be attributed to knowledge transfer.
To investigate whether the design of pretraining task itself 
plays a significant role
and can lead to further performance gains, 
we explored summarization datasets 
to prepare a battery of tasks 
found useful in creating summaries.
But these pretraining tasks performed comparably to
more generic pretraining tasks used in literature.
Our work suggests that understanding pretraining may have 
more to do with poorly-understood aspects
of how initialization influences optimization
than with knowledge transfer.

\bibliography{anthology,custom}
\bibliographystyle{acl_natbib}

\newpage


\appendix
\section{Appendix}
\label{sec:appendix}

\vspace{5px}
\noindent \textbf{Hyperparameters \quad} 
We use the T5-Small architecure with 60.5 million parameters as our transformer-based model.
The models are all trained using the BertAdam optimizer with a learning rate of $10^{-4}$.
For the pointer-generator model, the token embedding size is 128,
its encoder is a bidirectional LSTM with hidden size 256
the decoder is a unidirectional LSTM of the same size. 
The entire model had 4.4 million parameters.
For a fair comparision, we use wordpiece tokenization with all models with the same tokenizer
and vocabulary as used by the standard T5 model.
The validation metric used in all experiments was accuracy on the next-token prediction 
on the summaries. A patience value of 5 epochs was used for early stopping.

For CNN-Daiymail dataset, we truncated the input and output lengths 
according to \citet{zou2020pre} (Table~\ref{tab:hparams_lengths}).
We use the same lengths for the XSum dataset as well .
For the Rotowire and Rottentomatoes dataset, the input and output lengths were
much longer and even with a batch size of 1, we had to truncate them 
to values that allowed us to accommodate training with the available
GPU memory (32GB).
While decoding, we used beam search with beam size 4, and 
set the minimum and maximum decoding lengths  to the 
5 and 95 percentile of their observed distribution.

\vspace{5px}
\noindent \textbf{Computing infrastructure \quad} 
Most experiments were carried out on 8 Nvidia V100 GPUs with 32 GB of memory.
Some experiments with CNN-Dailymail and XSum datasets were carried out on 
4 Nvidia RTX2080Ti GPUs with 11GB of memory.

\vspace{5px}
\noindent \textbf{Exclusions from ensemble of our tasks \quad} 
When creating artificial summaries requires using multiple of our 
proposed elementary tasks, the different keywords added 
to the input signal to the model which tasks are 
required for it.
Three of our proposed tasks 
do not always involve keyword addition--- 
\emph{CopyFirstSentence, CopyLastSentence, CheckKeyword}.
Hence we exclude them when creating the pretraining corpus with 
our ensemble of tasks. We also exclude the \emph{SumOfNumbers}
and \emph{CompareNumbers} tasks because they could not be learnt
even in isolation by a randomly initialized T5 model 
training on 100k datapoints.

\vspace{5px}
\noindent \textbf{Details of dataset splits \quad} 
For the Rotowire and RottenTomatoes datasets, we use the standard
training, validation and test splits  with sizes shown in Table~\ref{tab:dataset_splits_sizes}.
For the CNN-Dailymail and XSum datasets, we use the standard test splits, but reduce the training and validation set sizes to 10k and 1k respectively by uniformly subsampling from the standard full dataset splits.

\vspace{5px}
\noindent \textbf{Evaluation metrics \quad} 
We measure the quality of generated summaries using 
ROUGE scores~\cite{lin2002manual} which measure n-gram overlap
between a generated and reference summary to assess its 
quality. We use the ROUGE-1,2 and L variants of this metric
which measure overlap in unigrams, bigrams and longest common
subsequence respectively. 
We also present the average performance of models at predicting 
the next token of a summary given all the ground truth past tokens (Table~\ref{tab:tokpred_results}).
To measure this, we use the accuracy and the negative-log-likelihood
metrics which are standard for multi-class classification. We average
these metrics across different decoding timesteps of summary generation,
and then average it again across all the summaries in the test set.

\begin{table*}[h!]
    \centering
\begin{tabular}{lcccc}
\toprule
{} &         \textbf{CNN-DM-10K} &      \textbf{XSum-10K} &   \textbf{RottenTomatoes}  &       \textbf{Rotowire} \\
\midrule
Train        & 10000 & 10000 & 2458  & 3398  \\
Validation   & 1000  & 1000  &  536  & 727   \\
Test         & 11490 & 11333 &  737  & 728   \\
\bottomrule

\end{tabular}
    \caption{Sizes for Train, validation and test splits for all datasets}
    \label{tab:dataset_splits_sizes}
\end{table*}

\begin{table*}[h!]
    \centering
\begin{tabular}{lcccc}
\toprule
{} &     \textbf{CNN-DM-10K} &      \textbf{XSum-10K} &       \textbf{RottenTomatoes}  &       \textbf{Rotowire}  \\
\midrule
max source length &  512 &  512 &    6000 &  5160 \\
max target length &  256 &  256 &  $\infty$ &   815 \\
batch size        &   16 &   16 &       1 &     1 \\
max decode length &  148 &   42 &      52 &   815  \\
min decode length &   44 &   18 &      16 &   223  \\
\bottomrule
\end{tabular}
    \caption{Hyperparameters used for fine-tuning models on the 4 datasets}
    \label{tab:hparams_lengths}
\end{table*}

\begin{table*}[t]
    \centering
    \begin{tabular}{l | c c | c c | c c | c c}
    \toprule
    \textbf{Experiment}     &  \multicolumn{2}{c}{\textbf{CNN-DM-10K}}  &  \multicolumn{2}{c}{\textbf{XSum-10K}}  &  \multicolumn{2}{c}{\textbf{Rottentomatoes}} &  \multicolumn{2}{c}{\textbf{Rotowire}}  \\
\toprule
{} &  Acc &  NLL &  Acc &  NLL &  Acc &  NLL &  Acc &  NLL  \\
\midrule
T5-OffShelf                  &           65.15 &        1.71 &          53.68 &       2.34 &            51.78 &         2.77 &          68.04 &       1.50 \\
T5-RandomInit                  &           29.78 &        4.92 &          32.60 &       4.75 &            24.75 &         5.36 &          48.30 &       2.61 \\
\midrule
\multicolumn{9}{c}{{Nonsense Upstream Corpus}}\\
\midrule
T5-OurTasks        &           54.74 &        3.18 &          38.98 &       4.27 &            33.42 &         5.08 &          63.59 &       1.78 \\
T5-STEPTasks           &           54.71 &        3.18 &          39.47 &       4.21 &            28.65 &         5.13 &          58.89 &       1.99 \\
\midrule
\multicolumn{9}{c}{{Real Upstream Corpus}}\\
\midrule
T5-OurTasks      &           54.87 &        2.93 &          41.21 &       3.76 &            39.64 &         4.12 &          64.02 &       1.78 \\

T5-STEPTasks     &           57.91 &        2.46 &          46.83 &       3.08 &            45.34 &         3.43 &          64.08 &       1.63 \\
\midrule
\multicolumn{9}{c}{{PG Models Randomly Initialized vs Pretrained (Nonsense Upstream Corpus)}}\\
\midrule
PG-RandomInit               &           51.14 &        2.91 &          33.05 &       4.14 &            33.35 &         4.37 &          59.12 &       1.92 \\
PG-OurTasks                 &           51.70 &        2.89 &          33.80 &       4.14 &            34.40 &         4.29 &          59.30 &       1.92 \\
PG-STEPTasks                &           51.79 &        2.88 &          34.13 &       4.14 &            35.06 &         4.21 &          59.00   &       1.94  \\
\bottomrule
\end{tabular}
    \caption{Accuracy (Acc) and negative log likelihood (NLL) for next token prediction on summaries}
    \label{tab:tokpred_results}
\end{table*}

\begin{table*}[t!]
    \centering
    \begin{tabular}{l|ccc|c}
    \toprule
    {Pretraining task} &    R1  & R2 & RL  & Pr\%\\
    \midrule
CopyKwdMultipleSent-Shuffled         &   \textbf{23.34} &    5.46 &   15.41  &  99.66 \\
TopicSegregation                     &   23.04 &    \textbf{7.79} &   \textbf{16.52}  &  99.88 \\
TruncateSentence                     &   17.07 &    2.50 &   11.81  &  100.00 \\
CopyQuoted                           &   11.03 &    1.32 &    8.32  &  99.82 \\
BreakClauses                         &   10.46 &    1.18 &    7.95  &  99.80 \\
CopyKwdMultipleSent-InOrder          &   10.14 &    1.14 &    7.70  &  99.84 \\
ReplaceClassKeyword                  &    9.70 &    0.95 &    7.36  &  99.98 \\
ParaphraseWords                      &    9.70 &    0.99 &    7.42  &  99.98 \\
CopyKwdOneSentence                   &    9.45 &    1.06 &    7.23  &  99.90 \\
CopyFirstSentence                    &    9.28 &    1.08 &    7.22  &  99.88 \\
CopyBulleted                         &    9.01 &    1.00 &    6.88  &  99.58 \\
CopyKwdMultipleSent-Sorted           &    8.48 &    0.83 &    6.59  &  99.68 \\
MajorityKeyword                      &    8.45 &    0.85 &    6.49  &  100.00 \\
ThresholdNumber                      &    7.83 &    0.77 &    6.05  &  100.00 \\
CheckKeyword                         &    7.79 &    0.77 &    5.94  &  100.00 \\
CopyLastSentence                     &    7.78 &    0.72 &    6.12  &  98.40 \\
JoinClauses                          &    7.72 &    0.81 &    6.09  &  98.82 \\
ClassifyKeyword                      &    6.80 &    0.62 &    5.34  &  100.00 \\
LargestNumber                        &    6.52 &    0.58 &    5.14  &  99.88 \\
SumOfNumbers                         &    5.03 &    0.40 &    4.14  &  25.06 \\
CompareNumbers                       &    1.89 &    0.04 &    1.75  &  48.88 \\
    \bottomrule
    \end{tabular}
    \caption{For different models pretrained on one individual task each, their performance on CNN-Dailymail-10K in terms of ROUGE (R1,R2,RL), and their accuracy in percentage on the pretraining task (Pr\%)}
    \label{tab:individual_task_effect_full}
\end{table*}

\begin{table*}[]
    \centering
    \resizebox{\linewidth}{!}{    
    \begin{tabular}{c|p{10cm}}
    \toprule
    \textbf{Elementary subtask} & \textbf{Description} \\
    \midrule
CheckKeyword &	Check if the input has a special keyword or not. \\
\hline
ClassifyKeyword &	Input contains 1 of 10 special keywords - 5 or them are positive and 5 of them are negative adjectives. Task is to tell whether mentioned adjective was positive or negative \\
\hline
MajorityKeyword &	Out of two given keywords, find which one occurs more number of times \\
\hline
CopyFirstSentence &	Copy first sentence \\
\hline
CopyBulleted &	Exactly one sentence is a bullet point and starts with the bullet marker. You have to copy over that sentence without copying the marker. \\
\hline
CopyQuoted &	Copy text within quotes \\
\hline
CopyLastSentence &	Copy last sentence \\
\hline
CopyKwdOneSent &	Copy single sentence containing one of many special defined keywords  \\
\hline
CopyKwdMultipleSentInOrder &	Copy all sentences containing any special keyword in the same order as they appear in text. \\
\hline
CopyKwdMultipleSentSorted &	Copy all sentences containing keywords but sort them according to the canonical ordering of keywords  \\
\hline
CopyKwdMultipleSentShuffled &	Copy all sentences containing keywords in any order. The sentences in ground truth may be any possible order.  \\
\hline
ReplaceClassKeyword &	There exist many keywords, each belonging to one of 3 classes. You have to mention  the class of the mentioned keyword \\
\hline
CompareNumbers &	Given two numbers in the text, say which one is bigger \\
\hline
SumOfNumbers &	Sum numbers \\
\hline
ThresholdNumber &	The input contains a number between 0 and 100. You have to say if the number was above or equal to the threshold of 50 of lower than it  \\
\hline
LargestNumber &	Find out largest of one or more numbers in the input. \\
\hline
TruncateSentence &	Copy a sentence but only till the cutoff keyword is encountered \\
\hline
BreakClauses &	Break a single sentence into multiple ones containing one clause each \\
\hline
JoinClauses &	Join clauses from multiple sentences to make one longer sentence \\
\hline
ParaphraseWords &	Copy the sentence containing one of pre-specified special keywords. But replace the keyword with any of its multiple synonyms. The $j^{th}$ synonym of $i^{th}$ keyword $src_i$ is given by $target_{ij}$ \\
\hline
TopicSegregation &	Copy all sentences containing keywords belonging to different classes but put them in corresponding sections (each class gets a separate section, which can be empty too, sections always occur in sorted order)  \\
\bottomrule
    \end{tabular}
    }
    \caption{21 extracted elementary summarization subtasks and their descriptions}
    \label{tab:21tasks_large}
\end{table*}

\begin{table*}[]
    \centering
    \begin{tabular}{lll}
    \toprule
    \textbf{Domain} & \textbf{Dataset name} & \textbf{Paper using the dataset} \\
    \midrule
\multirow{5}{*}{News} & CNN-Dailymail & \citet{see2017get} \\
    & NYT & \citet{paulus2018deep} \\
    & Gigaword & \citet{paulus2018deep} \\
    & XSUM & \citet{liu2019text} \\
    & Newsroom & \citet{zhang2020pegasus} \\
    \midrule
    \multirow{2}{*}{Code}	& Code to Documentation dataset & \citet{iyer2016summarizing} \\
    & Git diff to commit-message dataset & \citet{allamanis2016convolutional} \\
    \midrule
    \multirow{3}{*}{Scientific Paper} & Arxiv & \citet{cohan2018discourse} \\
    & Pubmed & \citet{cohan2018discourse} \\
    & ScisummNet & \citet{yasunaga2019scisummnet} \\
    \midrule
    Patent	& BigPatent & \citet{sharma2019bigpatent} \\
    \midrule
    Instructional guides &	Wikihow & \citet{zhang2020pegasus} \\
    \midrule
    Social media post & Reddit-TIFU & \citet{zhang2020pegasus}\\
    \midrule
    Email & AESLC & \citet{zhang2020pegasus} \\
    \midrule
    Bills & BillSum & \citet{zhang2020pegasus} \\
    \midrule
    \multirow{4}{*}{Reviews} & Amazon reviews & \citet{gerani2019modeling} \\
     & Yelp reviews & \citet{chu2019meansum}\\
     & CNET reviews & \citet{gerani2019modeling}\\
    \midrule
    \multirow{2}{*}{KeyValue Attributes} & Wikibio & \citet{lebret2016neural} \\
    & E2E dataset & \citet{novikova2017e2e} \\
    \midrule
    \multirow{4}{*}{Knowledge Graphs} & DBPedia triples to Wikipedia & \citet{vougiouklis2018neural} \\
    & AMR to sentence dataset & \citet{song2018graph} \\
    & Agenda & \citet{koncel2019text} \\
    & WebNLG & \citet{moryossef2019step} \\
    \midrule
    Numerical Table & Rotowire box-score & \citet{puduppully2019data} \\
    \midrule
    Miscellaneous webpages & Wikisum & \citet{liu2018generating} \\
    \midrule
    \multirow{2}{*}{Conversations} & SamSum & \citet{gliwa2019samsum} \\
    & AMI & \citet{wang2013domain} \\
    \bottomrule
    \end{tabular}
    \caption{Existing summarization datasets in various domains, along with corresponding papers that use them and came up during the search procedure to characterize elementary tasks in summarization}
    \label{tab:dasasets_domains}
\end{table*}

\end{document}